\documentclass[10pt,twocolumn,letterpaper]{article}

\usepackage{cvpr}              
\usepackage{verbatim}
\definecolor{cvprblue}{rgb}{0.21,0.49,0.74}
\usepackage[pagebackref,breaklinks,colorlinks,allcolors=cvprblue]{hyperref}

\def\paperID{51} 
\def\confName{CVPR}
\def\confYear{2025}

\title{EV-LayerSegNet: Self-supervised Motion Segmentation using Event Cameras}

\author{
    Youssef Farah\textsuperscript{1} \quad
    Federico Paredes-Vallés\textsuperscript{2} \quad
    Guido C.H.E. De Croon\textsuperscript{2} \quad \\
    Muhammad Ahmed Humais\textsuperscript{1} \quad
    Hussain Sajwani\textsuperscript{1} \quad
    Yahya Zweiri\textsuperscript{1} \quad \\
    \textsuperscript{1}Advanced Research and Innovation Center, Khalifa University \quad
    \textsuperscript{2}MAVLab, TU Delft \\
    {\tt\small \textsuperscript{1}\{youssef.farah, 100061899, hussain.sajwani, yahya.zweiri\}@ku.ac.ae} \\ 
    {\tt\small \textsuperscript{2}fedeparedesv@gmail.com, \textsuperscript{2}G.C.H.E.deCroon@tudelft.nl}
}
\makeatletter
\let\@oldmaketitle\@maketitle
\renewcommand{\@maketitle}{\@oldmaketitle
\centering
\setcounter{figure}{0}
  \includegraphics[width=0.91\textwidth]{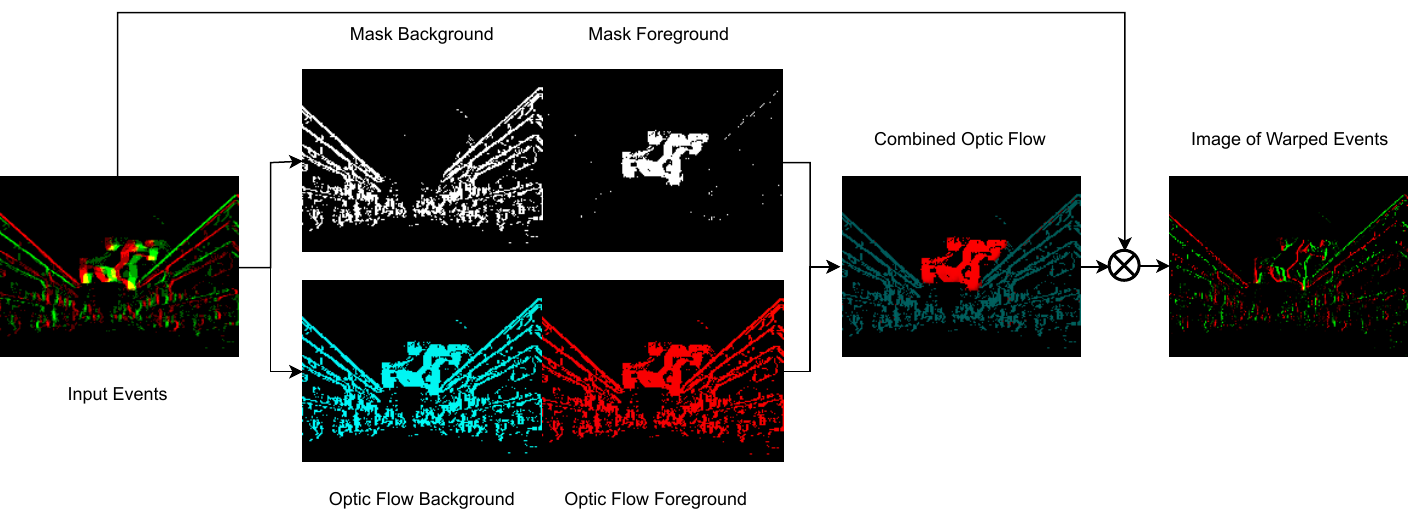}
  \captionof{figure}{Our method takes as input an event volume that results in a blurry scene. It then attempts to generate two different masks to differentiate the background from the foreground. Next, it estimates the affine optical flow for both models, and combines the flow together using the masks. Finally, it warps the events according to the combined flow. A successful motion deblur leads to accurate segmentation.}
}

\begin{document}
\maketitle
\begin{abstract}
Event cameras are novel bio-inspired sensors that capture motion dynamics with much higher temporal resolution than traditional cameras, since pixels react asynchronously to brightness changes. They are therefore better suited for tasks involving motion such as motion segmentation. However, training event-based networks still represents a difficult challenge, as obtaining ground truth is very expensive, error-prone and limited in frequency. \if Self-supervision is key in unlocking the true potential of event cameras on motion segmentation, but there is no existence of such learning-based methods. Current unsupervised methods in literature exploit events properties and use expectation maximization to jointly estimate the right clusters and associated motion model, without any learning thus preventing true application in the real world. To this end \fi In this article, we introduce EV-LayerSegNet, a self-supervised CNN for event-based motion segmentation. Inspired by a layered representation of the scene dynamics, we show that it is possible to learn affine optical flow and segmentation masks separately, and use them to deblur the input events. The deblurring quality is then measured and used as self-supervised learning loss. We train and test the network on a simulated dataset with only affine motion, achieving IoU and detection rate up to 71\% and 87\% respectively\footnote{Project page: \url{fulmen67.github.io/EV-LayerSegNet}}. 
\end{abstract}

\section{Introduction}

Event cameras, such as the Dynamic Vision Sensor (DVS), are novel bio-inspired sensors that perceive motion by detecting brightness changes instead of capturing intensity changes in images within a time interval~\cite{eventsurvey}. In other words, if at time $t$ a brightness change larger than the threshold $C$ is detected by a pixel with $x$ and $y$ coordinates, an event is generated by recording the brightness increase or decrease, the pixel location and the time the event was generated. In contrast, traditional cameras record color intensities in all pixels at a specified frame rate.

This fundamental shift in the acquisition of visual information brings several advantages. High temporal resolution and low latency (in the order of $\mu s$) make event cameras well suited for recording very fast moving scenes without suffering from motion blur as frame-based cameras. The High Dynamic Range (HDR) allows the cameras to capture details in very dark or very bright scenes, making them appropriate to be used in light challenging environments such as the underwater ocean, night driving or fireworks display~\cite{10.1109/TPAMI.2024.3496788}. 

The combination of these advantages unlocks the possibility of performing tasks inaccessible until now for frame-based cameras such as low-latency optical flow estimation, high-speed control and tracking, and Simultaneous Localization and Mapping (SLAM)~\cite{eventsurvey}. Nevertheless, a new generation of event-based vision algorithms needs to be developed to enable the event cameras to express their full potential, as the data structure of events is not compatible with the conventional algorithms based on images.

Furthermore, an additional challenge is posed by the lack of ground truth from real-word datasets, at microsecond resolution and with HDR. In image-based datasets (COCO~\cite{cocodataset}, BD100K~\cite{Yu2018BDD100KAD}), the ground truth is often obtained by humans manually annotating each frame. This process is already expensive and laborious for image-based datasets, and becomes unfeasible for event data, which are sparse and have low latency. To avoid manual annotation, additional sensors and cameras were used to generate the ground truth in traditional event-based datasets such as DSEC~\cite{9387069} and MVSEC~\cite{8288670}. Yet, sensors and cameras are limited by their natural Field Of View (FOV), spatial and temporal resolution. Therefore, it is of paramount importance to develop algorithms that are able to perform their tasks by relying on the nature of events, without the use of ground truth that are expensive to collect and that carry error measurements from additional sensors.

In the context of motion estimation and its use in other tasks such as surveillance, tracking or obstacle avoidance, segmenting the scene into independently moving objects is fundamental and often referred as \textit{motion segmentation}~\cite{9010722}. Even though great progress has been made using frame-based cameras, the latter are not ideally suited for tasks involving motion as they suffer from motion blur and keep track of all pixel values even when no motion occurs~\cite{9010722}. Under constant illumination, event cameras are well suited for this task, since events are sampled at exactly the same rate of the scene dynamics and the information acquired is only related to the motion of the camera and the objects in the scene.

Event-based motion segmentation can be broadly categorized in model-based methods \cite{9010722, 9561755, 9613780} and learning-based methods \cite{9157202,9196877,9636506}. Model-based methods exploit the nature of event data to cluster events generated by Independently Moving Objects (IMO) in an iterative fashion, thus they are unsupervised and do not use ground truth. Yet, the iterative approach requires long computational time and limits their real-word applicability. On the other hand, learning-based methods apply deep learning techniques while also relying on the event characteristics. In some instances, they perform motion segmentation in combination with other tasks such as tracking and optical flow, but they all require ground truth or pretrained weights during training. To the best of our knowledge, there is no existence of a learning-based method that is self-supervised, thus learning directly from the input events instead of relying on ground truth or pretrained weights from supervised methods. 

Looking outside of the event domain, significant progress has been made in self-supervised image segmentation. These methods segment the scene into the dominant moving object as foreground and the rest as background, and learning is driven by image or optic flow reconstruction. In most instances, they frequently make use of layered representation~\cite{334981}. The pixels are separated into layers based on motion similarity, and they are moved according to the associated motion and synthesized together to reconstruct the next image in the sequence. Recently, Shrestha \etal~\cite{9647227} integrated this process into a end-to-end differentiable CNN pipeline which segments the dominant moving object in a video using two consecutive frames.

Inspired by their work, we propose a self-supervised convolutional neural network for motion segmentation using events. Specifically, we transfer the CNN segmentation approach from~\cite{9647227} to the event domain, and combine it with the encoder-decoder structure from~\cite{8953979}. We show that, under the assumption of affine motion and constant brightness, it is possible to segment independently moving objects using contrast maximization loss.

We identify our main contributions as follows:

\begin{enumerate}
    \item We propose a novel self-supervised network for learning event-based motion segmentation by introducing a novel optical flow module that enables self-supervised learning of affine optical flow and a segmentation module that learns separately the masks corresponding to the independently moving objects.
    \item We contribute a new event-based dataset of several simulated backgrounds and objects moving according to affine motion.
    \item We demonstrate through experiments that our method shows superior performance in comparison with the state-of-the-art in unsupervised motion segmentation.  
\end{enumerate}

\section{Related Work}

In computer vision, motion segmentation is defined as the task of retrieving the shape of moving objects~\cite{8120119}. Image difference such as in~\cite{imagedifference} attempts to detect moving objects by finding intensity differences between pixels across video frames. Layer-based techniques \cite{layers,1541236} divide frames into layers, based on the number of uniform motions. Statistical methods such as Expectation Maximization (EM) are among the most common methods used. Recent developments in deep learning and optical flow estimation allowed a sharp improvement in this motion task, often combined with statistics and layer-based techniques. However, the main limitation of the state-of-the-art methods in motion segmentation is the reliance on ground truth labels, which limit the applicability of such methods scenes outside the annotated datasets. In this regard, attempts to develop self-supervised methods have resulted in frameworks developed in the frame domain, while methods for event cameras were yet to be found.

\subsection{Frame-based Motion Segmentation}

 In literature, many methods are considered unsupervised or self-supervised, but only during inference and test. In reality, parts of their architecture (i.e. networks, masks) are pretrained on ground truth. For example, COSNet~\cite{8953755} uses co-attention to capture rich correlations between frames of a video, but masks are pretrained on ground truth. Similarly, Ye \etal~\cite{Ye_2022_CVPR} perform motion segmentation using global sprites, but it also requires precomputed masks. Li \etal~\cite{8578781} propose the use of instance embeddings to find the moving object based on motion saliency and objectness. However, the dense embeddings are obtained from an instance segmentation that is performed by a network pretrained on static images. In contrast, we consider as self-supervised methods only the ones which do not need any type of ground truth during training and inference, and do not rely on pretrained weights from supervised methods.

    Early attempts of self-supervised or unsupervised learning of motion segmentation start from ConvNet~\cite{8100121}. The network learns high-level features in single frames as follows. A set of single frames with optical flow maps is passed to the network. It then generates the segmentation masks by associating optical flow vectors with similar direction and rate. Using these masks as pseudo-ground truth, the network tries to reproduce the masks by looking only at the static frames . 

Yang \etal~\cite{9711323} propose a network that performs foreground/background motion segmentation by using a precomputed optic flow as input. Slot attention is then used to group together pixels with visual homogeneity, but the result are very sensitive to the accuracy of the optical flow estimation.

While in  \cite{9647227,9711323} the models try to segment the moving object by using the optical flow of the object itself, Yang \etal~\cite{8954455} propose to find the moving object by using a Generator (G) and an Inpainter (I) network. The generator applies a mask to the input optic flow, with the aim of hiding the optic flow of the foreground. Next, the generator passes the masked optic flow to the inpainter, which has the task of reconstructing the optic flow that has been hidden by the generator. Accurate motion segmentation is achieved when the optic flow reconstruction works poorly. It obtains good result in public datasets, however the model fails to capture complete objects or differentiate regions in the background. This is because it operates with one single scale temporal information and it also introduces the bias from the camera motion. To solve this issue, Yang \etal~\cite{9707090} extend this work with MASNet, where they expand the network pipeline with more generator and inpainter modules to deal with multiple input optic flow maps.

Finally, Shrestha \etal~\cite{9647227} propose LayerSegNet, a network that combines a layered representation of the scene and affine optical flow estimation. The network takes two images as input and aims to separately estimate the motion models (foreground and background) and the masks. Subsequently, they apply the masks to the first image and warp the masks according to the associated motion models. They then combine the two masks together to generate the second image, and they use the difference between the generated second image and actual second image as learning parameter. 

\subsection{Event-based Motion Segmentation}

Event-based motion segmentation is very recent and attempts to make use of the outstanding properties of event cameras.

Stoffregen \etal~\cite{9010722} use Expectation Maximization and propose the idea of segmenting moving objects by simultaneously grouping the events into different clusters and estimating the motion model associated to each cluster. The events are then warped according to the associated motion model to produce an Image of Warped Events (IWE). An iteration process is then used to find the right motion models and clusters such that the objective function is maximized. The model is proven not to be sensitive to the number of clusters specified in the iteration process and works well in real-world dataset,but it is not able to estimate the number of moving objects in the scene. 

Zhou \etal~\cite{9613780} propose to solve this issue by introducing two spatial regularizers, which minimize the number of clusters. The underlying concept remains the same as in~\cite{9010722}, however the input events are first initialized in a space-time event graph cut and the clusters are smoothly sharped via the addition of energy terms in the objective function.

Always based on event clustering, Chethan \etal~\cite{9561755} split the scene into multiple motions and merge them, allowing also for feature tracking.

Despite these methods produce promising results, they are not appropriate to be deployed in real-world applications as the iteration processes takes considerable amount of time and computational effort, which inhibit the potential applications on mobile platforms such as drones. Therefore, a learning-based approach needs to be developed.

Sanket \etal~\cite{9196877} propose EVDodgeNet, a learning-based pipeline that solves simultaneously motion segmentation, optical flow and 3-D motion, but relies on ground truth masks. Instead, Mitrokhin \etal~\cite{9157202} use Graph Convolutional Networks to learn motion segmentation from a 3-D representation of events. Chethan \etal~\cite{9636506} also propose SpikeMS, the first Spiking Neural Network (SNN) for event-based motion segmentation. Very recently, Georgoulis \etal~\cite{10550654} performs the task by first compensating the events with egomotion, and then use attention modules for temporal consistency. Alkendi \etal~\cite{10.1109/TMM.2024.3521662} uses instead Graph Transformers neural networks to denoise the scene and perform segmentation.

Despite these efforts, the aforementioned methods continue to depend on ground truth, leaving a gap in the availability of self-supervised approaches. In this direction, Wang \etal~\cite{wang2023unevmosegunsupervisedeventbasedindependent} generate pseudo-labels from geometrical constraints, and Arja \etal~\cite{arja_motionseg_2024} propose the use of self-supervised vision transformers. Both works rely on pretrained weights from supervised networks.

Inspired by LayerSegNet~\cite{9647227}, we propose an end-to-end CNN architecture that learns motion segmentation by jointly estimating affine optical flow and segmentation masks, using contrast maximization as learning loss and without using any sort of pretrained weights.


\section{Method}
The approach to the segmentation task is inspired by~\cite{9647227}. However, given the unique nature of event data, a distinct approach to the input is needed. To this end, we use the input event representation from~\cite{8953979}. We then jointly estimate the segmentation masks and the affine optical flow maps. Next, we apply the segmentation masks to the associated optical flow, and we combine these maps to obtain a single optical flow map. We then use this map to warp the events forward and backward in time, and we apply the self-supervised loss as in~\cite{NEURIPS2021_39d4b545}. The more the image of warped events is deblurred with respect to the input events, the more accurate are the segmentation masks.

\subsection{Input Event Representation}

Selecting the right representation of events for a given task is still a challenging problem. Event-based methods can process single events $e_k \doteq (x_k, t_k, p_k)$, however events alone carry very little information and are subject to noise. It is often preferred to process a group of events $\mathcal{E} \doteq \{e_k\}^{N_e}_{k=1}$ that yields a sufficient signal-to-noise ratio, which also carries more information for the given task. The most common methods either discretize the group of events in different frames of event counts \cite{Lee2020SpikeFlowNetEO,9811821,9577656,Stoffregen2020ReducingTS,8953979} or the per-pixel average/most recent event timestamps \cite{9413238,9341224,Zhu2018EVFlowNetSO}. 

Since we develop a non-recurrent CNN pipeline, temporal information need to be encoded in the input, thus we adopt the event representation from~\cite{8953979}. Given $N$ input events and $B$ bins, we first scale the event timestamps in the range $[0, B - 1]$ and generate the event volume:

\begin{equation}
    t_i^*=(B-1)\left(t_i-t_0\right) /\left(t_N-t_1\right)
\end{equation}
\begin{equation}
  V(x, y, t)=\sum_i p_i k_b\left(x-x_i\right) k_b\left(y-y_i\right) k_b\left(t-t_i^*\right)  
\end{equation}
\begin{equation}
    k_b(a)=\max (0,1-|a|)
\end{equation}

In order to perform 2D convolutions, the time domain is treated as a channel in a traditional 2D image.

\subsection{Motion Model}

The authors in~\cite{9647227} propose the segmentation of the dominant moving object in the scene using a two-layer representation. The foreground layer captures the dominant object, while the background layer represents the rest of the scene.

In our case, assuming constant illumination, events are generated by the relative motion between the camera and the scene. Since events are triggered by brightness changes, these are produced by moving edges. Hence, if the camera is moving and the scene is static, the events are generated by the motion of the camera. When objects are also moving within the scene, the generation of events occurs due to both the edges of moving objects and the edges of static objects caused by the egomotion of the camera.

Let us assume a scene where an object and background are moving distinctively. We also assume that the motion is affine, hence the scene may undergo translation, rotation, shearing and scaling. The two motions can then be described by two motion models $A_1$ and $A_2$:

\begin{figure*}[t]
    \centering
    \includegraphics[width = \linewidth]{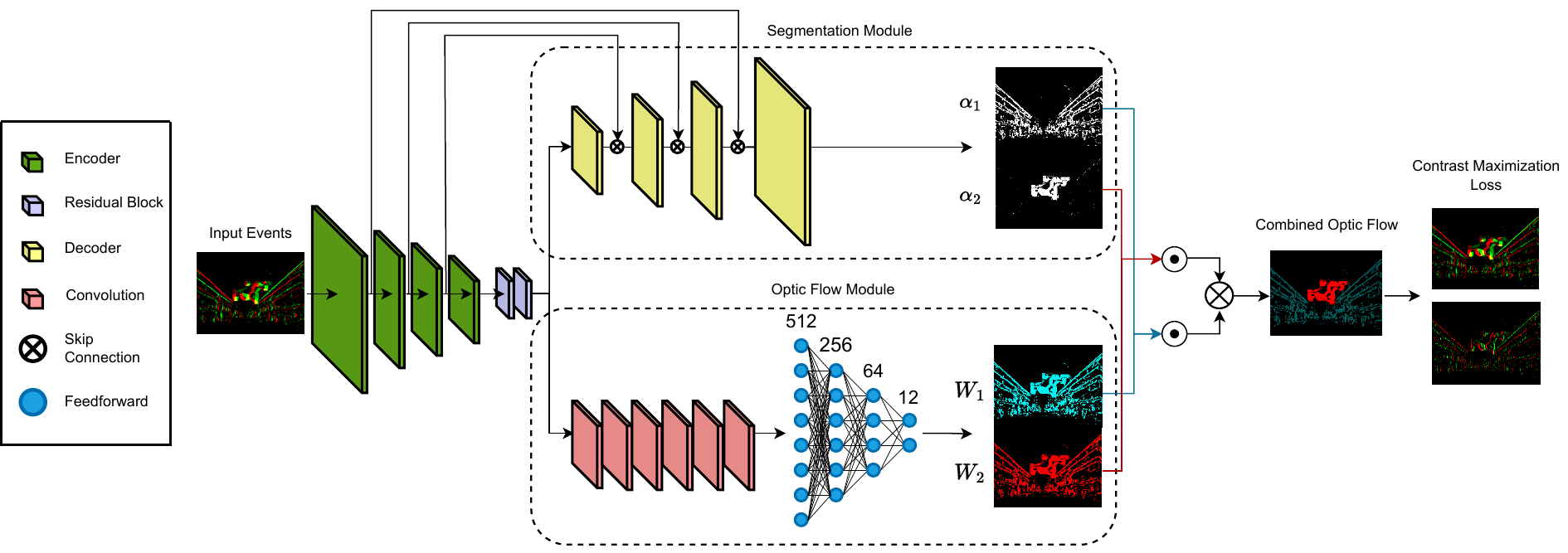}
    \caption{EV-LayerSegNet architecture. Events are downsampled by 4 encoding layers before being passed to 2 residual blocks. The output of the residual blocks is then passed to the segmentation and optical flow module. The segmentation module upsamples with 4 decoder layers connected with the encoders by skip connections. The optical flow module is composed of 6 convolutions and a feedforward network of 4 layers. }
    \label{fig: network}
\end{figure*}

\begin{equation}\label{eq: 6}
    W_{i}(x, y)=\mathbf{A}_{i}
    \begin{bmatrix}
    1 \\
    x \\
    y
    \end{bmatrix}
    =
    \begin{bmatrix}
    a_{i}^{1} & a_{i}^{2} & a_{i}^{3} \\
    a_{i}^{4} & a_{i}^{5} & a_{i}^{6}
    \end{bmatrix}
    \begin{bmatrix}
    1 \\
    x \\
    y
    \end{bmatrix}, \quad i \in\{1,2\}
\end{equation}

where $W_i$ represent the dense flow maps corresponding to affine motion matrix $A_i$. 

Corresponding to the two affine motions, we separately generate two alpha masks, which represent the moving objects. Ideally, the segmentation masks should be a one-hot vector for each pixel, which assigns the pixel to the corresponding object. However, such operation would be non differentiable, preventing the backpropagation of the gradients during training. We overcome this issue by assigning two arbitrary numbers for each pixel, where each number corresponds to one layer. We then apply softmax to ensure that the values are bounded in [0,1] and the values sum up to 1. For each pixel, we then retain the maximum value in the two layers, and set the other one to zero. This operation, called maxout operation, makes the classification differentiable. It also makes sure that one pixel can not belong to two objects at the same time.

Next, we compute the combined optical flow by element-wise multiplication of the affine flow maps with the corresponding alpha maps:

\begin{equation}\label{eq: 8}
    W_{\text{comb}}=\alpha_{1} \odot W_{1} + \alpha_{2} \odot W_{2}
\end{equation}

The maxout operation ensures that each pixel will have flow components from only one motion model, scaled with the respective value of the alpha map to which the pixel is associated. During training, we expect that the network will adjust itself to the scaling effect of the flow computation. 

\subsection{Self-supervision by Contrast Maximization}

Self-supervised learning is achieved by applying contrast maximization~\cite{8954483}. As described in~\cite{NEURIPS2021_39d4b545}, events which are generated by the same moving edge encode exact optical flow. Since these events are misaglined in the input partition (motion blur), the events can be propagated to a reference time $t_{\text{ref}}$ using per-pixel optical flow $\boldsymbol{u}(\boldsymbol{x})=(u(\boldsymbol{x}), v(\boldsymbol{x}))^T$ to realign them and show the initial edge that has generated the events:

\begin{equation}
    \boldsymbol{x}_i^{\prime}=\boldsymbol{x}_i+\left(t_{\mathrm{ref}}-t_i\right) \boldsymbol{u}\left(\boldsymbol{x}_i\right)
\end{equation}

The metric used to measure the deblurring quality is the the per-pixel and per-polarity average timestamp of the image of warped events \cite{8593805,8953979}. The lower the loss, the better the deblurring, which consequently means that the estimated optical flow and alpha maps are more accurate. 

We initially generate an image of the average timestamp at each pixel for each polarity using bilinear interpolation, as in~\cite{NEURIPS2021_39d4b545}:

\begin{equation}\label{eq:52}
\begin{array}{r}
T_{p^{\prime}}\left(\boldsymbol{x} ; \boldsymbol{u} \mid t_{\mathrm{ref}}\right)=\frac{\sum_{j} \kappa\left(x-x_{j}^{\prime}\right) \kappa\left(y-y_{j}^{\prime}\right) t_{j}}{\sum_{j} \kappa\left(x-x_{j}^{\prime}\right) \kappa\left(y-y_{j}^{\prime}\right)+\epsilon} \\

j=\left\{i \mid p_{i}=p^{\prime}\right\}, \quad p^{\prime} \in\{+,-\}, \quad \epsilon \approx 0
\end{array}
\end{equation}

The loss is then the sum of the squared temporal images. This is also scaled by the sum of pixels with at least one event, in order to prevent the network from keeping events with large timestamps out of the image space, such that they would not contribute to the loss function. 

\begin{equation}\label{eq:53}
\mathcal{L}_{\text {contrast }}\left(t_{\mathrm{ref}}\right)=\frac{\sum_{\boldsymbol{x}} T_{\pm}\left(\boldsymbol{x} ; \boldsymbol{u} \mid t_{\mathrm{ref}}\right)^{2}}{\sum_{\boldsymbol{x}}\left[n\left(\boldsymbol{x}^{\prime}\right)>0\right]+\epsilon}
\end{equation}

The warping process is performed both forward ($t_{\text{ref}}^{fw}$) and backward ($t_{\text{ref}}^{bw}$) to prevent temporal scaling issues during backpropagation. The total loss is then the sum of the backward and forward warping loss, and $\lambda$ is a scalar balancing the Charbonnier smoothness prior $\mathcal{L}_{\text{smooth}}$~\cite{413553}.

\begin{equation}\label{eq:54}
    \mathcal{L}_{\text {contrast }} =\mathcal{L}_{\text {contrast }}\left(t_{\text {ref }}^{\mathrm{fw}}\right)+\mathcal{L}_{\text {contrast }}\left(t_{\text {ref }}^{\mathrm{bw}}\right) 
\end{equation}

\begin{equation}
    \mathcal{L}_{\text {flow }} =\mathcal{L}_{\text {contrast }}+\lambda \mathcal{L}_{\text {smooth }}
\end{equation}

Notice that the self-supervised loss becomes a strong supervisory signal only when there is enough blur in the input event partition. It is of paramount importance to check that the input partition contains enough events for linear blur.
\section{Network Implementation}

Our network EV-LayerSegNet is similar to encoder-decoder architectures and is inspired by \cite{9647227,8953979}. 

As in~\cite{8953979}, events are downsampled by 4 encoder layers and passed to 2 residual blocks. We then stack the output of the residual blocks and pass it to the segmentation module and optical flow module.

\subsection{Optical Flow Module}
The optical flow module contains 6 convolutional layers, each followed by leaky ReLU activation. We then flatten the output of the last convolution layer and pass it to a feedforward network consisting of 4 layers (512, 256, 64 and 12 output units), followed by tanh activation except the last layer. We then use the output of the feedforward network  and split it to two sets of 6 affine motion parameters, and we compute the two flow maps $W_1$ and $W_2$.

\subsection{Segmentation Module}
In the segmentation part, the output of the residual blocks is bilinearly upsampled by 4 decoding layers. Each decoding layer is connected to the respective encoding layer by skip connection. We apply a leaky double-rectified ReLU activation (\textit{leaky DoReLU})~\cite{9647227} which promises an increase in segmentation performance:

\begin{equation}
y(x)= \begin{cases}1+\frac{x-1}{\gamma} & , x>1 \\ x & , \text { if } 0 \leq x \leq 1 \\ \frac{x}{\gamma} & , x<0\end{cases}
\end{equation}

In this module, each decoding layer is followed by leaky DoReLU activation with $\gamma$ = 100. At the last layer, softmax is applied instead to ensure that the channel values are bounded [0,1] and sum up to 1.

\section{Experiments}
\subsection{Training Details}

We implement our pipeline in Pytorch. We use Adam optimizer~\cite{Kingma2014AdamAM} and a learning rate of $1 \cdot 10^{-5}$ . At training, we used a batch size of 8 and train for 400 epochs. The input consists of N = 200,000 events, which is a sufficient number of events to obtain blur in the scene. The Charbonnier loss is balanced with weight $\lambda$ = 0.001 .

\subsection{Dataset}\label{subsec: dataset}

Due to the absence of a sufficiently large dataset with objects moving with affine motion, we train EV-LayerSegNet on our dataset, generated using ESIM~\cite{Rebecq18corl}. Specifically, we make use of the Multiple 2D Objects rendering engine to generate the simulated events. We use similar settings as in a DVS camera, with resolution 640x480 pixels and threshold C = 0.5. For the training set, we generate 1000 sequences of 1 second each, using 10 background images and 10 foreground objects. Similarly, the test set is composed of 25 sequences of 1 second each, with five background images and five foreground images. The ground truth is generated by using a threshold on optical flow, since the foreground objects move significantly faster than the background images.

\subsection{Evaluation Metrics}

We use \textit{Intersection over Union} (IoU) and \textit{Detection Rate} (DR) as standard metrics for our quantitative evaluation~\cite{9613780}. IoU is formulated as in \cref{eq: iou}, where \textit{$S_D$} stands for the predicted mask and \textit{$S_G$} for the ground truth mask:

\begin{equation}
\text{IoU} = \frac{S_D \cap S_G}{S_D \cup S_G}
\label{eq: iou}
\end{equation}

Detection rate returns 1 if the following conditions in \cref{eq: dr} are met, where \textit{$B_D$} and \textit{$B_G$} are the bounding boxes for predicted mask and ground truth:

\begin{equation}
B_D \cap B_G > 0.5 \quad \text{and} \quad (B_D \cap B_G) > (S_D \cap \overline{B_G})
\label{eq: dr}
\end{equation}

\section{Results}

\begin{figure*}[t]
    \centering
    \includegraphics[width = \linewidth]{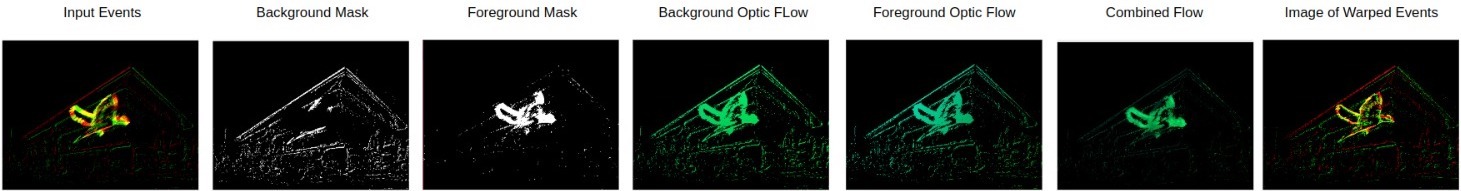}
    \caption{Test result example on synthetic dataset ("Bird in front of building" in \cref{tab: test results 50k}). The color scheme can be found in~\cite{NEURIPS2021_39d4b545}.}
    \label{fig: test results}
\end{figure*}

\subsection{Qualitative Evaluation}

\begin{figure}
    \centering
    \includegraphics[width=\linewidth]{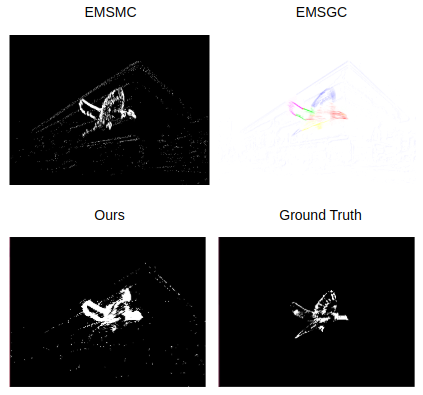}
    \caption{Qualitative comparison of test results from EMSMC~\cite{9010722}, EMSGC~\cite{9613780} and our method against ground truth ("Bird in front of building" in \cref{tab: test results 50k}).}
    \label{fig: comparison with ground truth}
\end{figure}

We provide a qualitative example in \cref{fig: test results}. The network effectively deblurs the scene, composed of a bird flying in the same direction of the camera but at a higher speed, as encoded by the colors of the optical flow maps. This shows that our method is also capable of segmenting challenging scenarios in which the background and foreground might appear to have similar motions, which could lead the network to group them together. \cref{fig: comparison with ground truth} shows the same scene compared to ground truth and the respective motion segmentation estimated by EMSMC~\cite{9010722} and EMSGC~\cite{9613780}. Our method correctly clusters the bird in the right class, but is subject to noise, as noticed from the sparse events of the background. In this case, EMSMC figuratively shows a better segmentation, while EMSGC segments the bird in different clusters. In the quantitative evaluation, although EMSGC mislabels the foreground in different clusters, we consider all clusters minus the background as one main cluster.

\subsection{Quantitative Evaluation}

We benchmark our method against EMSMC~\cite{9010722} and EMSGC~\cite{9613780} using mean intersection over union (Mean IoU) and mean detection rate (Mean DR) on our test dataset. This approach is relevant since both methods do not require any sort of ground truth or pretrained weights to perform in testing. The benchmark with EMSMC is fair since both methods require an initial number of clusters, which in our case is two (background and foreground). We compare results also with EMSGC, while keeping in mind the increased challenge of solving also for the increased number of clusters. Due to the high motion present in the test dataset, we set N = 50,000 for all methods. This choice was also motivated due to the extensive computational time needed for both EMSMC and EMSGC to perform. All methods have found challenging the test dataset, having relevant performances in 6 sequences shown in \cref{tab: test results 50k}. In most cases, our method demonstrates higher performance in segmenting the scene. In addition to this, our method performs on-the-fly motion segmentation, requiring few milliseconds to get the masks, while EMSMC and EMSGC, due to their iterations, require several minutes. 

\begin{table*}[ht]
\centering
\resizebox{0.7\textwidth}{!}{
\begin{tabular}{@{}lccccccc@{}}
\toprule
                              & \multicolumn{3}{c}{\textbf{Mean IoU}} & \multicolumn{3}{c}{\textbf{Mean Detection Rate}} \\ \midrule
\textbf{Sequence name}        & \textbf{EMSMC} & \textbf{EMSGC} & \textbf{Ours} & \textbf{EMSMC} & \textbf{EMSGC} & \textbf{Ours} \\
Drone above playground        & 0.20          & 0.05          & 0.71          & 0.00          & 0.00          & 0.87 \\
Plane over city               & 0.06          & 0.03          & 0.67          & 0.00          & 0.00          & 0.77 \\
Bird above playground         & 0.26          & 0.00          & 0.48          & 0.00          & 0.00          & 0.71 \\
Second drone above playground & 0.36          & 0.02          & 0.52          & 0.00          & 0.00          & 0.78 \\
Bird in front of building     & 0.20          & 0.41          & 0.10          & 0.00          & 0.21          & 0.00 \\
Helicopter over city          & 0.07          & 0.00          & 0.55          & 0.00          & 0.00          & 0.83 \\ \bottomrule
\end{tabular}}
\caption{Comparison between our method, EMSMC~\cite{9010722} and EMSGC~\cite{9613780} on 6 simulated test sequences using an identity camera matrix.}
\label{tab: test results 50k}
\end{table*}

\section{Ablation Study}\label{sec: ablation study}

\subsection{Leaky ReLU vs Leaky DoReLU}

In~\cite{9647227} the authors discovered that using leaky DoReLU activation function instead of leaky ReLU improves segmentation. The key distinction between these two activation functions lies in the restriction of the slope for input values above 1 in leaky DoReLU. The authors argue that the reason for the segmentation improvement with leaky DoReLU is the enhanced stability of gradients, considering their alpha maps are limited to the range [0,1].

Inspired by their segmentation approach, we chose to implement leaky DoReLU instead of leaky ReLU. Nevertheless, we investigated the significance of this modified activation function in our specific case.

For this reason, we created an alternative EV-LayerSegNet architecture where leaky ReLU activations are used instead of leaky DoReLU. We trained it on our dataset and subsequently tested it.

The results in \cref{tab: test results leakyrelu} show that the network encounters more difficulties compared to the baseline. This observation suggests that the leaky DoReLU activation improves our network's ability to achieve successful segmentation.

\begin{table}
\centering
\resizebox{\columnwidth}{!}{
\begin{tabular}{@{}lcccc@{}}
\toprule
                              & \multicolumn{2}{c}{\textbf{Mean IoU}} & \multicolumn{2}{c}{\textbf{Mean Detection Rate}} \\ \midrule
\textbf{Sequence name}        & \textbf{LeakyReLU}     & \textbf{Ours}    & \textbf{LeakyReLU}          & \textbf{Ours}          \\
Drone above playground        & 0.38               & 0.71             & 0.00                    & 0.87                   \\
Plane over city               & 0.48               & 0.67             & 0.65                    & 0.77                   \\
Bird above playground         & 0.31               & 0.48             & 0.00                    & 0.71                   \\
Second drone above playground & 0.48               & 0.52             & 0.78                    & 0.78                   \\
Helicopter over city          & 0.45               & 0.55             & 0.61                    & 0.83                   \\ \bottomrule
\end{tabular}}
\caption{Test results of our method with Leaky ReLU against baseline.}
\label{tab: test results leakyrelu}
\end{table}

\begin{figure}
    \centering
    \includegraphics[width = \columnwidth]{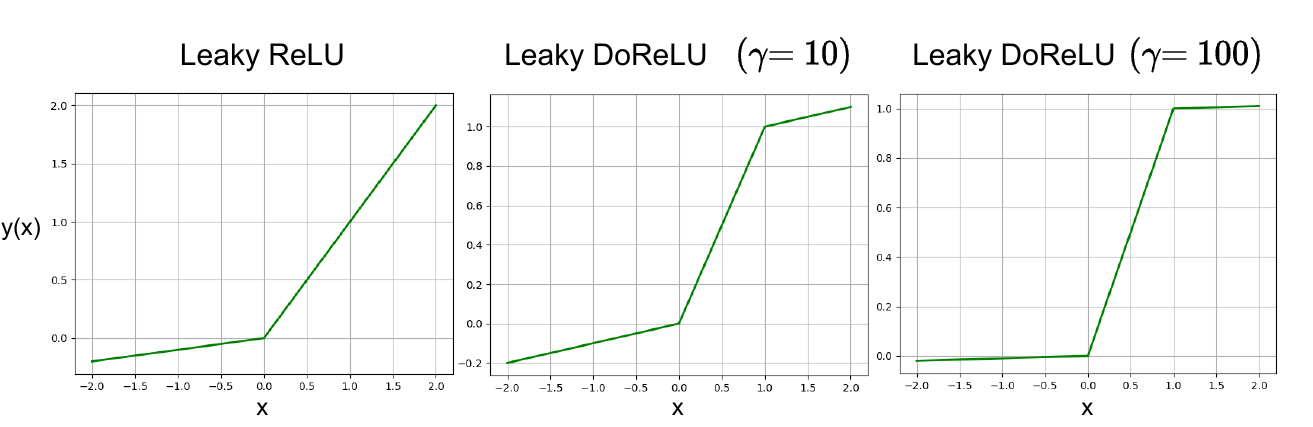}
    \caption{Visualization of leaky ReLU (left) and leaky DoReLU with slope coefficient $\gamma$ = 10  (center) and $\gamma$ = 100 (right). The x-axis is the input $x$ and the y-axis is the output $y(x)$. }
    \label{fig: functions}
\end{figure}

\subsection{Decreasing the slope in Leaky DoReLU}
By establishing the importance of the leaky DoReLU activation function in our segmentation module, we turned our attention to exploring the impact of modifying the slope coefficient $\gamma$. We investigated how the variation $\gamma$ would affect the segmentation performance of our network.

The characteristics of the activation functions involved are shown in \cref{fig: functions}. It can be observed that the leaky DoReLU exhibits a smaller slope compared to the leaky ReLU beyond the range of [0, 1]. This reduced slope plays a vital role in stabilizing the flow of gradients. We suspect that this is due to the function being more non-linear, improving our segmentation performance. To test this hypothesis, we decide to train the network with a higher slope ($\gamma$ = 10), therefore expecting poorer results.


The corresponding test results are presented in \cref{tab: test results gamma10}, where the baseline method performs significantly better. This suggests that a nearly flat slope outside of range [0, 1] in the leaky DoReLU function plays a vital role for segmentation.

\begin{table}
\centering
\resizebox{\columnwidth}{!}{
\begin{tabular}{@{}lcccc@{}}
\toprule
                              & \multicolumn{2}{c}{\textbf{Mean IoU}} & \multicolumn{2}{c}{\textbf{Mean Detection Rate}} \\ \midrule
\textbf{Sequence name}        & \textbf{$\gamma$ = 10}     & \textbf{Ours}    & \textbf{$\gamma$ = 10}          & \textbf{Ours}          \\
Drone above playground        & 0.06               & 0.71             & 0.00                    & 0.87                   \\
Plane over city               & 0.03               & 0.67             & 0.00                    & 0.77                   \\
Bird above playground         & 0.07               & 0.48             & 0.00                    & 0.71                   \\
Second drone above playground & 0.02               & 0.52             & 0.00                    & 0.78                   \\
Helicopter over city          & 0.21               & 0.55             & 0.00                    & 0.83                   \\ \bottomrule
\end{tabular}}
\caption{Test results of our method with $\gamma$ = 10 against baseline. Segmentation performance significantly deteriorates across all scenes.}
\label{tab: test results gamma10}
\end{table}

\section{Conclusion}

We present a novel end-to-end CNN network that performs self-supervised motion segmentation using event cameras. We take inspiration from the state-of-the-art methods in unsupervised motion segmentation, and use the latest development in event-based optical flow estimation using artificial neural networks. We show that it is possible to learn motion models and segmentation masks separately from a blurry scene, and combine them together to deblur the input events. To train and test our method, we generate a new dataset with background and objects moving affinely. We show that the network is able to correctly segment the background and the independently moving objects in the event stream. We also show that our method outperforms the state-of-the-art in unsupervised motion segmentation. On the other hand, our method is limited to affine motion and is sensitive to very high blurriness and noise. For future work, we seek to extend the training and testing dataset to have more diversity, and benchmark the method against other self-supervised methods that require pretrained weights, and finally supervised methods. In the longer term, we seek to test on a 3D real-world dataset. Additional recommendations would be to improved the optical flow module such that also non-affine motions are estimated, and develop the method using Spiking Neural Networks, therefore allowing for fast and low-power computation on board of weight-constrained vehicles such as drones.
We hope that our work draws attention to the potential that self-supervised learning can unlock for event-based cameras, allowing learning-based methods to be trained and used without expensive ground-truth annotations.

\section*{Acknowledgments}

This work is supported by Sandooq Al Watan under Grant SWARD-S22-015, STRATA Manufacturing PJSC, and
Advanced Research and Innovation Center (ARIC), which is jointly funded by Aerospace Holding Company LLC, a
wholly-owned subsidiary of Mubadala Investment Company PJSC and Khalifa University for Science and Technology.
 
{
    \small
    \bibliographystyle{ieeenat_fullname}
    \bibliography{main}
}


\end{document}